\documentclass[10pt,twocolumn,letterpaper]{article}

\usepackage{cvpr}
\usepackage{mathptmx}
\usepackage{graphicx}
\usepackage{graphicx}
\usepackage{amsmath}
\usepackage{amssymb}
\usepackage{blindtext}
\usepackage{siunitx}
\DeclareSIUnit\px{px}
\usepackage{tikz}
\usepackage{pgfplots}
\usepgfplotslibrary{polar,groupplots,dateplot}
\usetikzlibrary{patterns,shapes.arrows,positioning,decorations.pathreplacing}
\pgfplotsset{compat=newest}
\usepackage{pgfplotstable}
\usepackage{booktabs}
\usepackage{pifont}% http://ctan.org/pkg/pifont
\newcommand{\cmark}{\ding{51}}%
\newcommand{\xmark}{\ding{55}}%
\usepackage{makecell}
\usepackage{placeins}
\usepackage{tkz-kiviat}
\usetikzlibrary{shapes, arrows}
\usepackage{tikzscale}
\usepackage[upright]{fourier} 
\usepackage{tkz-kiviat,numprint,fullpage} 
\usetikzlibrary{arrows}
\usepackage{subfigure}

\usepackage{todonotes}

\usepackage[breaklinks=true,bookmarks=false]{hyperref}
\usepackage[nameinlink,capitalize]{cleveref}
\crefformat{equation}{(#2#1#3)}

\cvprfinalcopy % *** Uncomment this line for the final submission

\def\cvprPaperID{****} % *** Enter the CVPR Paper ID here

\begin{document}

%%%%%%%%% TITLE
\title{Cityscapes 3D: Dataset and Benchmark for 9 DoF Vehicle Detection}

% version of \and with smaller spacing
\makeatletter
\def\ands{%                  % \begin{tabular}
    \end{tabular}%
    \hskip 4pt \@plus.017fil%
    \begin{tabular}[t]{c}}%   % \end{tabular}
\makeatother

\author{Nils G{\"a}hlert$^{1,2}$
    \ands Nicolas Jourdan$^{1,3}$
    \ands Marius Cordts$^{1}$
    \ands Uwe Franke$^{1}$
    \ands Joachim Denzler$^{2}$\\[.6em]
    \ands $^{1}$Mercedes-Benz AG
    \ands $^{2}$University of Jena
    \ands $^{3}$TU Darmstadt
}

\maketitle
%\thispagestyle{empty}

%%%%%%%%% ABSTRACT
\begin{abstract}
    Detecting vehicles and representing their position and orientation in the three
    dimensional space is a key technology for autonomous driving.
    Recently, methods for 3D vehicle detection solely based on monocular RGB images gained popularity.
    In order to facilitate this task as well as to compare and drive state-of-the-art methods,
    several new datasets and benchmarks have been published.
    Ground truth annotations of vehicles are usually obtained using
    lidar point clouds, which often induces errors due to imperfect calibration or
    synchronization between both sensors.

    To this end, we propose \emph{Cityscapes 3D}, extending the original Cityscapes dataset with
    3D bounding box annotations for all types of vehicles. In contrast to existing
    datasets, our 3D annotations were labeled using stereo RGB images only and capture all
    nine degrees of freedom. This leads to a pixel-accurate reprojection in the RGB image and a higher
    range of annotations compared to lidar-based approaches. In order to ease
    multitask learning, we provide a pairing of 2D instance segments with 3D bounding boxes.
    In addition, we complement the Cityscapes benchmark suite with 3D vehicle detection based
    on the new annotations as well as metrics presented in this work.
    Dataset and benchmark are available online\footnote{\url{https://www.cityscapes-dataset.com/}}.
\end{abstract}

%%%%%%%%% BODY TEXT
% !TeX root = ../egpaper_final.tex

\section{Introduction}
3D object detection is commonly approached by using lidar or radar sensors as they
have exceptional physical properties for this task. More recently,
vision-based methods have been proposed that detect objects in 3D solely relying on
a single monocular RGB image \cite{brazil2019m3d,
    ku2019monocular,bao2019monofenet,manhardt2019roi,liu2019deep,li2019gs3d,ma2019accurate}.
These methods gained more and more interest as cameras are more prevalent than laser scanners
or radars and allow for a finer-grained classification.
Furthermore, they can be used as a redundant sensor in safety-critical applications such as autonomous driving.
As no explicit 3D information is encoded in RGB images, accurate depth
prediction is more challenging compared to lidar-based methods. This results in a
significant gap in detection performance, which becomes evident via
benchmarks that feature both modalities, \eg KITTI (best
lidar: \SI{79.71}{\percent} \cite{yang2019std}, best monocular:
\SI{10.74}{\percent} \cite{ma2019accurate}; 3D Average Precision (AP) for
category \emph{Car}) or nuScenes (best lidar \num{0.484} \cite{ye2020sarpnet},
best monocular: \num{0.384} \cite{simonelli2019disentangling}; nuScenes Detection
Score).

\begin{figure}[t]
    \begin{center}
        \includegraphics[width=\linewidth]{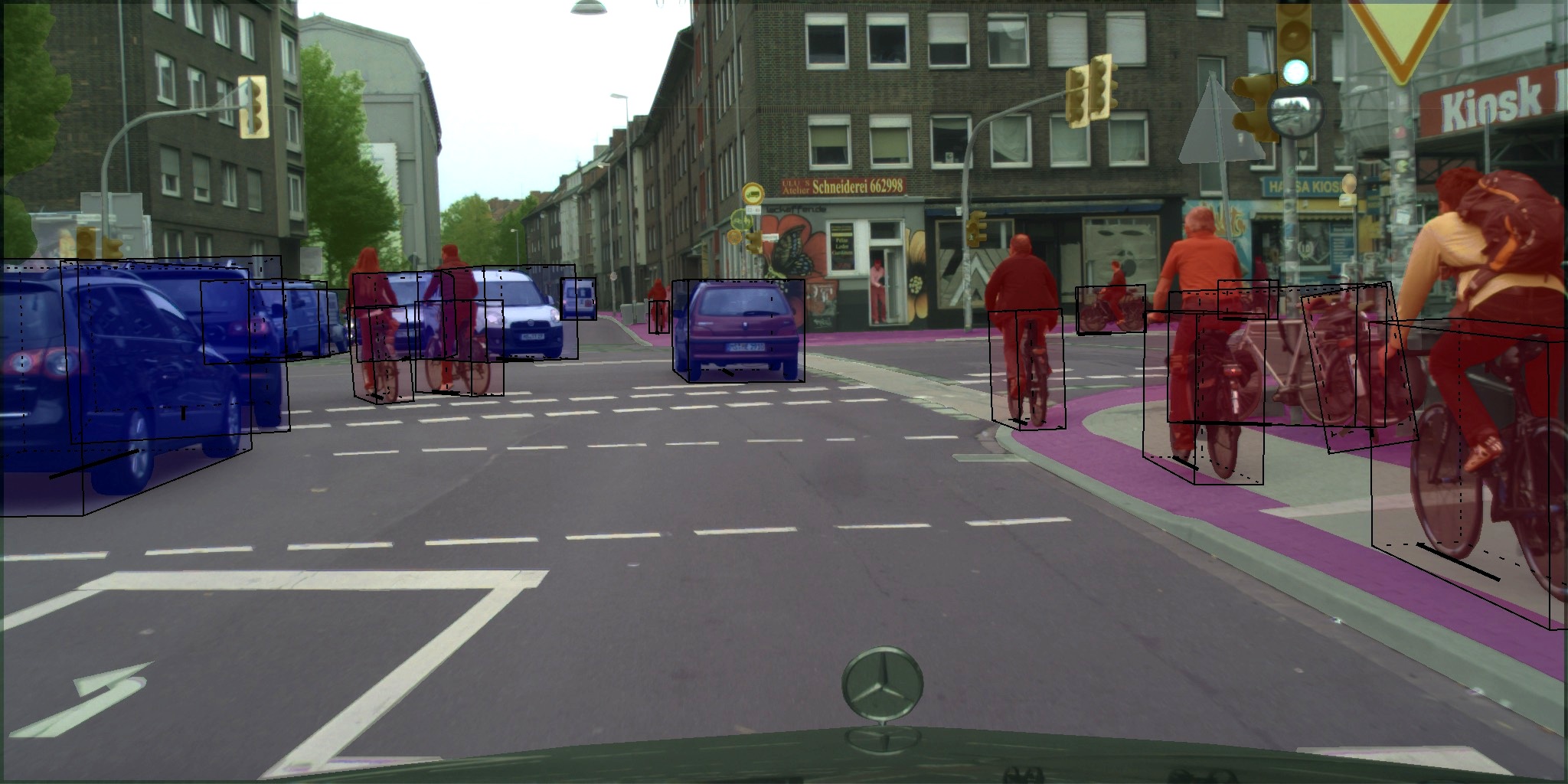}
    \end{center}
    \caption{Example image with 3D bounding boxes for vehicles. The box annotations feature a full 3D orientation including yaw, pitch and roll labels.\label{fig:spider}}
\end{figure}

To facilitate further research on monocular 3D object detection, we provide high
quality 3D bounding box annotations of all types of vehicles for the Cityscapes
dataset \cite{cordts2016cityscapes}, which is one of the most popular
datasets for semantic, instance, and panoptic segmentation. Based on these
annotations, we offer a benchmark for 3D vehicle detection such that
researchers can easily compare novel approaches with the state-of-the-art.
Dataset and benchmark complement \cite{cordts2016cityscapes} and will also
allow for future research, \eg joint 3D detection and instance segmentation. To this end,
we ensured consistency with the existing annotations and provide matches of 2D
instance ground truth masks with the new 3D bounding box annotations.

In contrast to existing 3D object detection datasets and benchmarks, Cityscapes
3D is especially tailored for monocular 3D objection detection, due to three
major design choices.

First, ground truth annotations are obtained using stereo RGB imagery only, which
overcomes limitations of other datasets. Related real-world datasets, \eg
\cite{Argoverse,geiger2012we,nuscenes2019,waymo_open_dataset},
commonly rely on lidar point clouds for 3D bounding box annotation.
This approach either requires high efforts regarding sensor setup, calibration and synchronization or suffers from
drawbacks due to
mismatches between both sensors, \cf \cref{fig:synchronization_error}. These
effects are most prominent in case of fast moving objects close to the
ego-vehicle, \ie exactly those objects that are highly relevant for self-driving
cars.

\begin{figure}[t]
    \includegraphics[width=\columnwidth]{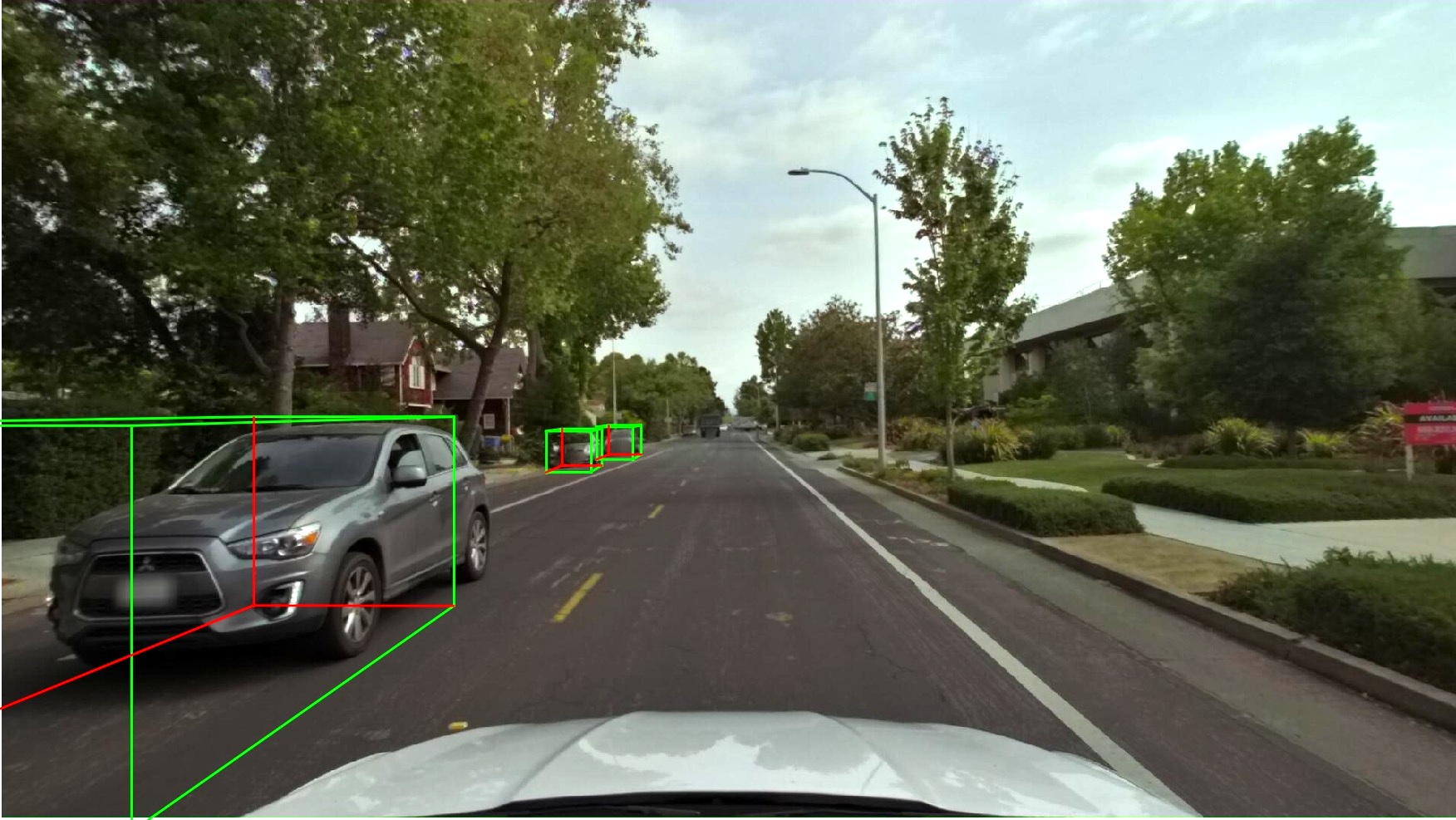}
    \vspace{-3mm}

    \includegraphics[width=\columnwidth, trim={0 0 0 176}, clip]{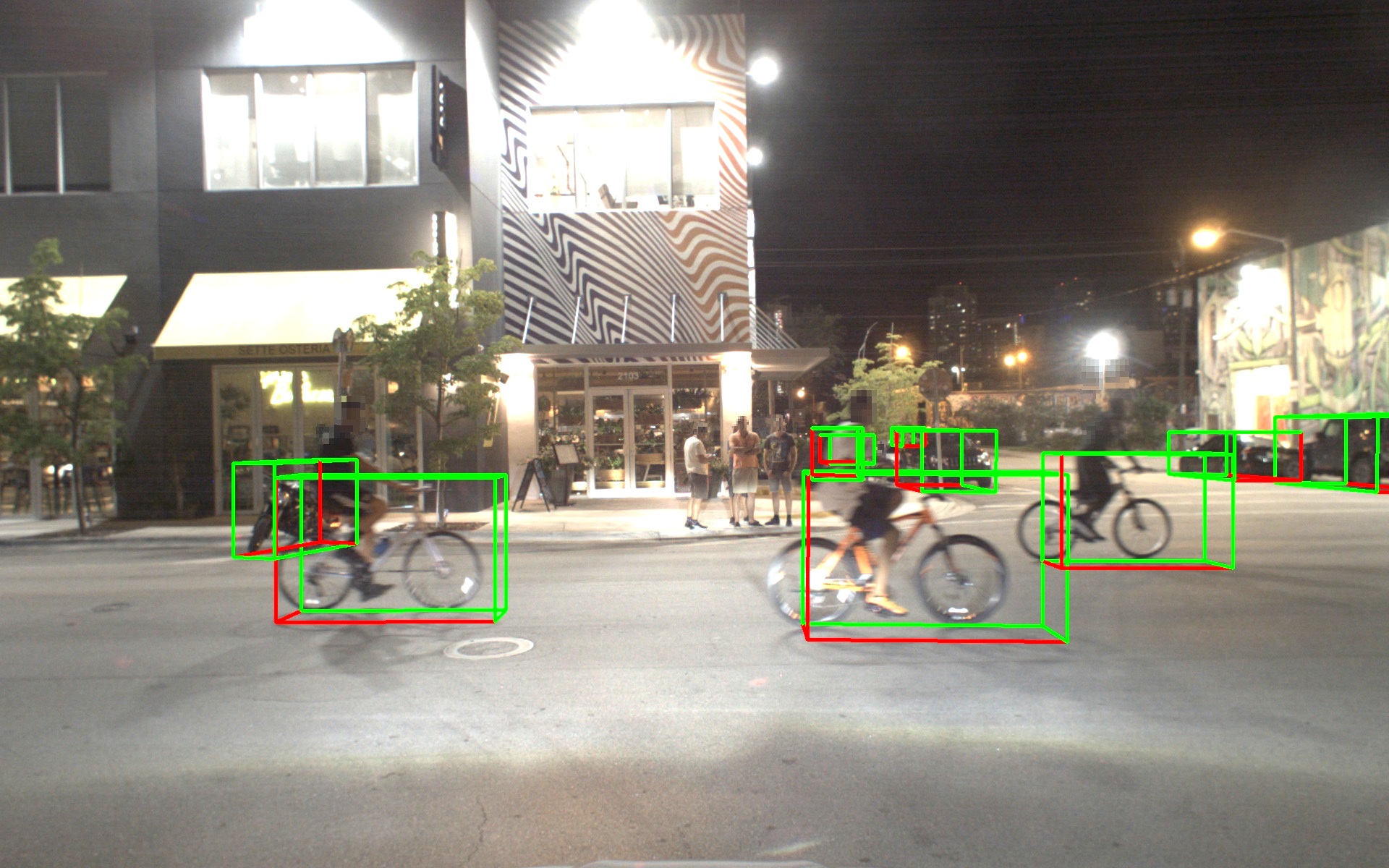}
    \caption{Examples of annotation artifacts due to lidar synchronization errors (\textbf{top}  \cite{lyft2019}, \textbf{bottom} \cite{Argoverse}). \label{fig:synchronization_error}}
\end{figure}

Second, we provide full 3D orientation annotations including yaw, pitch, and
roll angles to cover all nine degrees of freedom of a rigid object (position,
extent, and orientation). As slanted roads occur in real-world scenes, such a
representation is crucial to precisely describe and recognize vehicles in all
constellations, \cf \cref{fig:pitchroll}.

\begin{figure}[t]
    \includegraphics[width=\columnwidth]{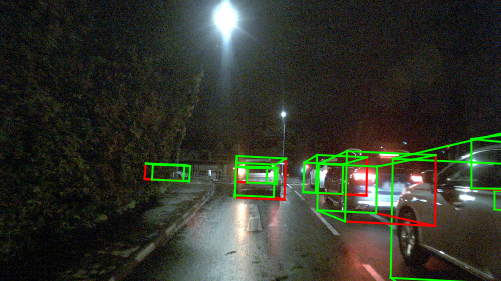}
    \vspace{-3mm}

    \includegraphics[width=\linewidth]{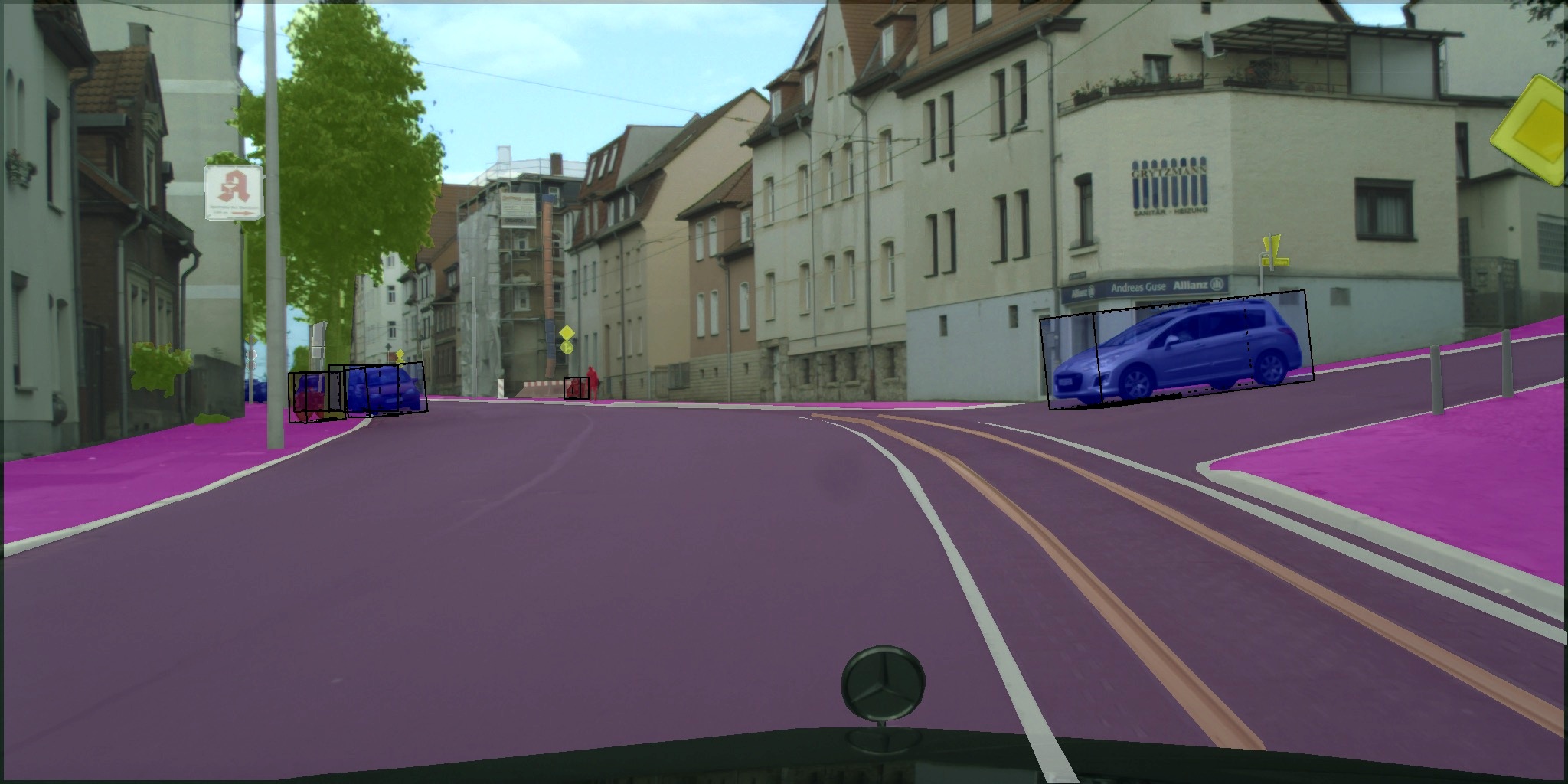}
    \caption{\textbf{Top}: Imprecise 3D box reprojections in \cite{nuscenes2019} due to missing pitch and roll annotations.  \textbf{Bottom}: Example for a car crossing from a steep street in Cityscapes 3D with \mbox{pitch $> 0$}. \label{fig:pitchroll}}
\end{figure}

Third, the fundamental difference between lidar and RGB data as an input modality
for 3D object detection is not sufficiently addressed in the current benchmark
methodologies. While precise depth information is inherently included in lidar data,
accurate depth estimation from monocular images is a challenging task. However, current benchmarks
and their employed metrics are primarily designed for lidar-based approaches and
often use the 3D Intersection over Union (IoU) metric with a high threshold.
Doing so effectively requires centimeter accuracy, which can barely be achieved by
vision-based approaches for distant objects and leads to a significant drop in accuracy compared to
lidar-based methods.
We therefore introduce novel metrics to assess the performance of
monocular 3D object detection.  Notably, the proposed metrics explicitly
evaluate the performance depending on the distance of the object to the
ego-vehicle.

% !TeX root = ../egpaper_final.tex

\section{Related Work}
% !TeX root = ../egpaper_final.tex

\begin{table*}[t]
  \caption{Overview on datasets for autonomous driving featuring image-based vehicle recognition via 2D instance segmentation or 3D bounding box detection. Cityscapes 3D is the only dataset based on real-world data that supports both tasks with paired 2D instance segmentation masks and 3D bounding boxes. Furthermore, Cityscapes 3D was labeled using stereo images only, overcoming shortcomings due to calibration and synchronization issues.\label{tab:datasets}\vspace{2pt}}
  \label{tab:dataset}
  \centering
  \begin{tabular}{lrccccc}
    \toprule
                                                               &                           & \multicolumn{3}{c}{Annotations}                           &                                                                           \\ \cmidrule{3-6}
    Name                                                       & Resolution                & Instance Masks                                            & 3D Boxes                  & 3D Based on & Paired & 3D Benchmark           \\ \midrule
    \textbf{Cityscapes 3D} + \cite{cordts2016cityscapes}       & \SI{2.1}{MP}              & \cmark                                                    & \cmark                    & stereo      & \cmark & \cmark                 \\[3pt]
    A*3D \cite{pham20193d}                                     & \SI{3.1}{MP}              & \xmark                                                    & \cmark                    & lidar       & \xmark & \xmark                 \\
    A2D2 \cite{aev2019}                                        & \SI{2.3}{MP}              & \cmark\footnotemark[1]                                    & \cmark                    & lidar       & \xmark & \xmark                 \\
    ApolloScape \cite{huang2018apolloscape}                    & \SI{9.2}{MP}              & \cmark                                                    & \cmark\footnotemark[2]    & lidar       & \xmark & \cmark                 \\
    Argoverse \cite{Argoverse}                                 & \SI{5.1}{MP}              & \xmark                                                    & \cmark                    & lidar       & \xmark & closed                 \\
    BDD100k \cite{yu2018bdd100k}                               & \SI{0.9}{MP}              & \cmark\footnotemark[2]                                    & \xmark                    & --          & \xmark & \xmark                 \\
    Boxy \cite{boxy2019}                                       & \SI{5.1}{MP}              & \xmark                                                    & trapezoid\footnotemark[1] & monocular   & \xmark & \cmark\footnotemark[3] \\
    KITTI \cite{geiger2012we}                                  & \SI{0.5}{MP}              & \cmark\footnotemark[1]                                    & \cmark                    & lidar       & \xmark & \cmark                 \\
    Lyft \cite{lyft2019}                                       & \num{1.3} \& \SI{2.1}{MP} & \xmark                                                    & \cmark                    & lidar       & \xmark & closed                 \\
    Mapillary Vistas \cite{neuhold2017mapillary}               & various                   & \cmark                                                    & \xmark                    & --          & \xmark & \xmark                 \\
    nuScenes \cite{nuscenes2019}                               & \SI{1.4}{MP}              & \cmark\footnotemark[2]\textsuperscript{,}\footnotemark[3] & \cmark                    & lidar       & \xmark & \cmark                 \\
    % Pandaset by Hesai
    Waymo Open \cite{waymo_open_dataset}                       & \SI{2.5}{MP}              & \xmark                                                    & \cmark                    & lidar       & \xmark & \cmark                 \\
    \midrule
    Synscapes \cite{wrenninge2018synscapes}                    & \SI{2.1}{MP}              & \cmark                                                    & \cmark                    & synthetic   & \cmark & \xmark                 \\
    Synthia \cite{ros2016synthia}                              & \num{0.9} \& \SI{2.1}{MP} & \cmark                                                    & \cmark                    & synthetic   & \cmark & \xmark                 \\
    VIPER \cite{richter2017playing}                            & \SI{2.1}{MP}              & \cmark                                                    & \cmark                    & synthetic   & \cmark & \xmark                 \\
    Virtual KITTI 2 \cite{gaidon2016virtual, cabon2020vkitti2} & \SI{0.5}{MP}              & \cmark                                                    & \cmark                    & synthetic   & \cmark & \xmark                 \\
    \bottomrule
    \multicolumn{5}{l}{\footnotesize\footnotemark[1] only some instances per semantic class in an image}                                                                                                                           \\[-3pt]
    \multicolumn{5}{l}{\footnotesize\footnotemark[2] only a subset of the images in the dataset}                                                                                                                                   \\[-3pt]
    \multicolumn{5}{l}{\footnotesize\footnotemark[3] to be released}                                                                                                                                                               \\
  \end{tabular}
\end{table*}

There are multiple datasets
available that address different problems in perception for autonomous driving
such as object detection or instance segmentation. An overview of current
datasets for autonomous driving with a focus on segmenting and detecting all types
of vehicles is given in \cref{tab:datasets}.

These datasets provide a wide range of driving scenarios as they were recorded
at various locations all over the world; \eg in the US \cite{Argoverse,nuscenes2019,waymo_open_dataset},
Singapore \cite{nuscenes2019,pham20193d}, China \cite{huang2018apolloscape},
or Germany \cite{geiger2012we,cordts2016cityscapes}.
Mapillary Vistas \cite{neuhold2017mapillary} even contains images from 6 continents.
While only half of the datasets include instance masks as ground truth annotations,
nearly all provide 3D bounding boxes. While instance masks are often only available
for a subset of the dataset, 3D bounding boxes are usually provided
for the whole dataset.

The majority of 3D bounding box annotations is labeled using lidar data \cite{waymo_open_dataset,nuscenes2019,lyft2019}.
However, using lidar data is vulnerable to calibration and synchronization
errors, \cf \cref{fig:synchronization_error} which may result in imprecise reprojections
into the RGB images. The only exceptions are
Boxy \cite{boxy2019}, which was annotated in the image domain to obtain
3D ground truth data, and ApolloScapes \cite{huang2018apolloscape} where CAD models were
used for ground truth generation. In this work, we build our annotation workflow
on top of stereo image pairs in order to obtain accurate and well-aligned 3D bounding boxes.

As shown previously,
bounding boxes that are generated from instance segmentation masks help to boost amodal 2D
object detection \cite{gaehlert2019vg} and can be beneficial for 3D object
detection from monocular RGB images as well \cite{li2019gs3d,manhardt2019roi}.
Furthermore, such paired annotations, \ie direct mappings between instance masks
and 3D bounding boxes, ease multitask learning. However,
Cityscapes 3D is the only dataset out of those in \cref{tab:datasets} that provides such a mapping
between the two modalities for all annotated images.

In addition to real-world datasets, there exist several synthetic ones such as \cite{wrenninge2018synscapes,ros2016synthia,richter2017playing,cabon2020vkitti2}.
By design, synthetic datasets include high quality labels for all types
of annotations. However, these datasets suffer from a domain gap towards real-world scenes,
which makes them less suitable for benchmark purposes.

% !Tex root = ../egpaper_final.tex
\section{Labeling Process}
\label{sec:labeling}
%\placeholder

Recent datasets for autonomous driving often include 3D bounding box annotations
that were mainly labeled in lidar point clouds, as can be seen in
\cref{tab:dataset}. Using these annotations to benchmark monocular 3D bounding
box detection methods proves difficult as the correctness of the projection into
corresponding RGB images relies heavily on correct cross-sensor calibration and
synchronization. This circumstance is highlighted in
\cref{fig:synchronization_error}.

In contrast, for Cityscapes 3D we aim at monocular 3D bounding
box detection. Thus, all 3D bounding box annotations are exclusively labeled
using the stereo camera, preventing issues with calibration and synchronization.
To this end, we exploit stereo data from \cite{cordts2016cityscapes} that was generated using semi-global matching (SGM) \cite{hirschmuller2007stereo}
with on-site calibration prior to each recording session.
The main challenge of labeling 3D bounding boxes in RGB images is the
indistinctness between depth and size of 3D objects in images. Objects of
vastly different size may look equally large in images when placed at
appropriate distances.
To overcome this issue, we employed two techniques during our annotation
workflow, \ie \emph{stereo point clouds} and \emph{size prototypes}.

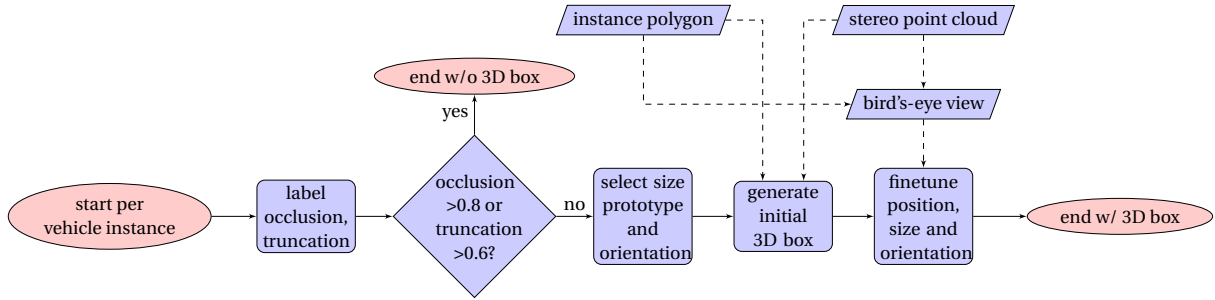
\begin{figure*}
    \centering
    \makebox[0pt]{%
        \resizebox{\textwidth}{!}{
            % Define block styles
\tikzstyle{decision} = [diamond, draw, fill=blue!20, 
    text width=4.5em, text badly centered, node distance=3cm, inner sep=0pt]
\tikzstyle{block} = [rectangle, draw, fill=blue!20, 
    text width=5em, text centered, rounded corners, minimum height=4em, node distance=3cm]
\tikzstyle{line} = [draw, -latex']
\tikzstyle{cloud} = [draw, ellipse,fill=red!20, node distance=3cm,
    minimum height=2em]
 \tikzstyle{root} = [rectangle, rounded corners, draw, fill=red!20, node distance=3cm]
\tikzstyle{data} = [draw, trapezium, 
                     trapezium left angle=70, trapezium right angle=110,
                     fill=blue!20, node distance=2cm]

\begin{tikzpicture}[auto]
    % Place nodes
    \node [cloud, align=center] (start) {start per \\ vehicle instance};
    \node [block, right of=start, node distance=3.5cm] (truncation) {label occlusion, \mbox{truncation}};
    \node [decision, right of=truncation] (occluded) {occlusion \textgreater 0.8 or \mbox{truncation}
    \textgreater 0.6?};
    \node [cloud, above of=occluded, node distance=2.5cm] (endoccluded) {end w/o 3D box};
    \node [block, right of=occluded] (selectprototype) {select size prototype and \mbox{orientation}};
    \node [block, right of=selectprototype, node distance=2.5cm] (sizeprototype) {generate initial 3D box};
    \node [block, right of=sizeprototype, node distance=2.5cm] (finetune) {finetune position, size and \mbox{orientation}};
    \node [data, above of=finetune, node distance=2cm] (bev) {bird's-eye view};
    \node [data, above of=bev, node distance= 1.5cm] (stereo) {stereo point cloud};
 	\node [data, left=2cm of stereo] (instance) {instance polygon};
    \node [cloud, right of=finetune, node distance=3.5cm] (endfinished) {end w/ 3D box};
    % Draw edges
    \path [line] (start) -- (truncation);
    \path [line] (truncation) -- (occluded);
    \path [line] (occluded) -- node {yes} (endoccluded);
    \path [line] (occluded) -- node {no} (selectprototype);
    \path [line] (selectprototype) -- (sizeprototype);
    \path [line, dashed] (stereo) -| (sizeprototype.60);
    \path [line, dashed] (instance) -| (sizeprototype.120);
    \path [line, dashed] (instance) |- (bev);
    \path [line, dashed] (stereo) -- (bev);
    \path [line, dashed] (bev) -- (finetune);
    \path [line] (sizeprototype) -- (finetune);
    \path [line] (finetune) -- (endfinished);
\end{tikzpicture}
        }
    }
    \caption{Workflow of creating the 3D bounding box annotations given the existing instance polygons and stereo measurements.\label{fig:flowchart}}
\end{figure*}

The labeling process of Cityscapes 3D is schematically visualized in \cref{fig:flowchart}.
We initialize our workflow with the vehicle instances as annotated in \cite{cordts2016cityscapes}.
For each vehicle, the occlusion as well as the truncation of the shown object
are manually estimated in \SI{10}{\percent} intervals. Vehicles that
are more than \SI{80}{\percent} occluded or \SI{60}{\percent} truncated are filtered out,
\ie they are not annotated with a 3D bounding box and also set to \emph{ignore} in the Cityscapes 3D benchmark, \cf \cref{sec:benchmark}.
For all remaining vehicle instances, the labeler selects a finer-grained vehicle type,
\cf \cref{tab:prototypes} for a list of available categories.
The selected category is then used as size prototype,
\ie an initial size as well as an initial orientation of the 3D
bounding box annotation are assigned.
The initial position of the 3D bounding box is then determined by the stereo
measurements contained in the instance-level annotation polygon from \cite{cordts2016cityscapes}. Initial
dimensions paired with an initial position estimate prevent errors caused by the
trade-off between object size and depth of three dimensional objects in images.
Preliminary experiments showed that this procedure significantly improves the annotation quality
and speed compared with labeling from scratch.

\begin{table}
    \caption{Size prototypes used for initial dimensions of the 3D bounding box annotations during labeling.}
    \begin{tabular}{ l c c c }
        \toprule
                               & \multicolumn{3}{c}{Dimensions [\si{\meter}]}                  \\
        Prototype Name         & Height                             & Width & Length \\
        \midrule
        Mini Car               & 1.45                               & 1.65  & 2.70   \\
        Small Car              & 1.45                               & 1.65  & 4.00   \\
        Compact Car            & 1.45                               & 1.80  & 4.30   \\
        Sedan                  & 1.45                               & 1.81  & 4.70   \\
        Station Wagon          & 1.50                               & 1.85  & 4.90   \\
        Box Wagon              & 1.80                               & 1.80  & 4.35   \\
        Sports Utility Vehicle & 1.70                               & 1.90  & 4.70   \\
        Pick-Up                & 1.80                               & 1.92  & 5.30   \\
        Sports Car             & 1.30                               & 1.81  & 4.13   \\
        \midrule
        Small Van              & 1.90                               & 1.90  & 5.40   \\
        Large Van              & 2.60                               & 1.85  & 6.50   \\
        Caravan                & 3.00                               & 2.20  & 7.20   \\
        \midrule
        Mini Truck             & 3.00                               & 2.20  & 7.00   \\
        Small Truck            & 3.45                               & 2.32  & 7.95   \\
        Medium Truck           & 4.00                               & 2.50  & 12.00  \\
        Large Truck            & 4.00                               & 2.55  & 6.80   \\
        Truck Trailer          & 4.00                               & 2.55  & 13.60  \\
        \midrule
        Urban Bus (Solo)       & 3.10                               & 2.55  & 12.00  \\
        Urban Bus (Front)      & 3.10                               & 2.55  & 7.40   \\
        Urban Bus (Back)       & 3.10                               & 2.55  & 7.40   \\
        Coach Bus              & 3.80                               & 2.55  & 14.00  \\
        \midrule
        Bicycle                & 1.10                               & 0.42  & 1.80   \\
        Motorbike              & 1.12                               & 0.80  & 2.20   \\
        \bottomrule
    \end{tabular}
    \label{tab:prototypes}
\end{table}

The labeler is subsequently asked to fine-tune orientation, position and dimensions of
the 3D bounding box in a bird's-eye view projection of the stereo measurements
while checking the plausibility of the annotation in the RGB image. The RGB
image information enables the labeling of full 3D orientation information (yaw,
pitch, roll). An example of an RGB image with corresponding bird's-eye view is shown in \cref{fig:rgb_labeling}.

\begin{figure*}[t]
    \centering
    \subfigure[3D bounding box annotation projected into the RGB image]{
        \includegraphics[height=4cm]{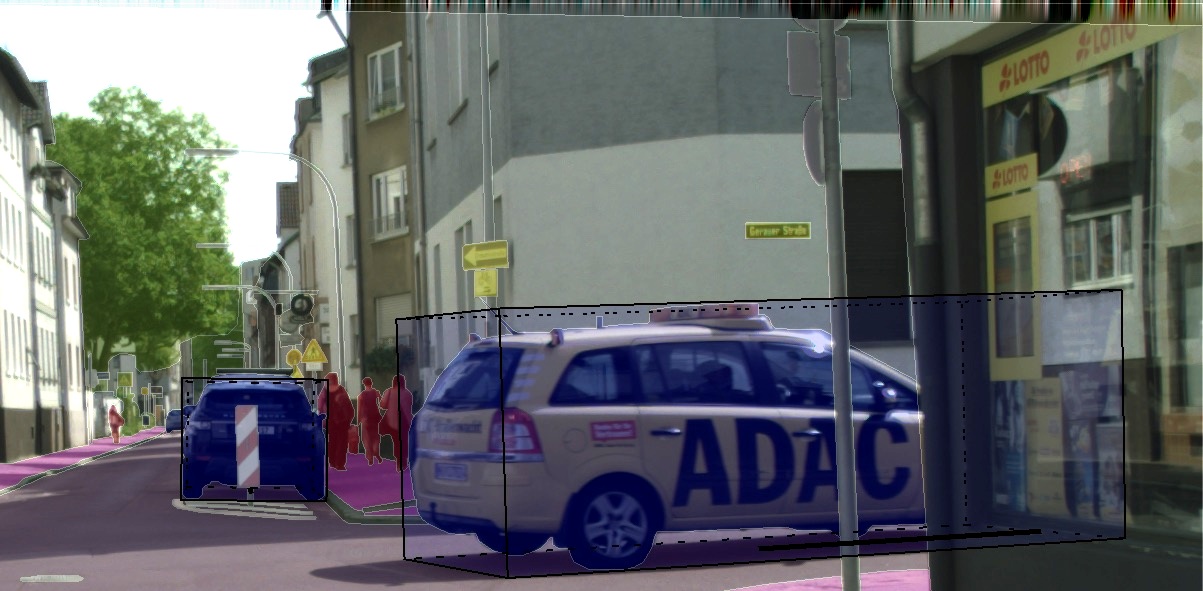}
    }
    \subfigure[Corresponding bird's-eye view labeling aid]{
        \includegraphics[height=4cm] {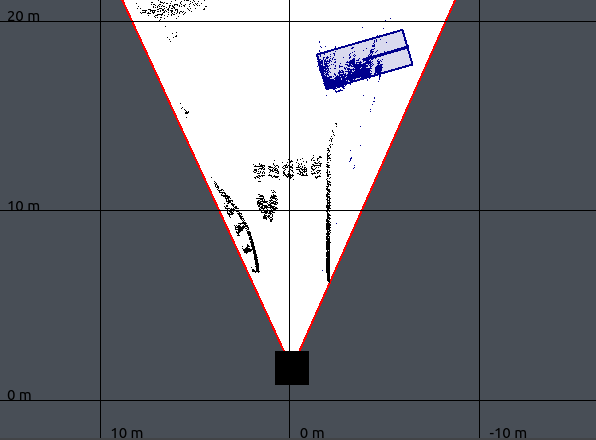}
    }
    \caption{Example for bird's-eye view labeling aid. 3D bounding box annotations and the stereo point cloud filtered for vehicles and markings are shown from a top view perspective of the scene.\label{fig:rgb_labeling}}
\end{figure*}

In case of a vehicle with several moving parts, \eg an articulated bus,
multiple 3D bounding boxes were annotated to cover each movable part.
\cref{fig:bus} depicts an example of a bended articulated bus.
As in some situations, \eg very crowded scenes, it is not possible to
identify single object instances, the whole object group is marked as such
and ignored during evaluation, \cf \cref{sec:benchmark}. An example for a crowded scene is shown in
\cref{fig:ignore}.
\begin{figure}[t]
    \includegraphics[width=\linewidth]{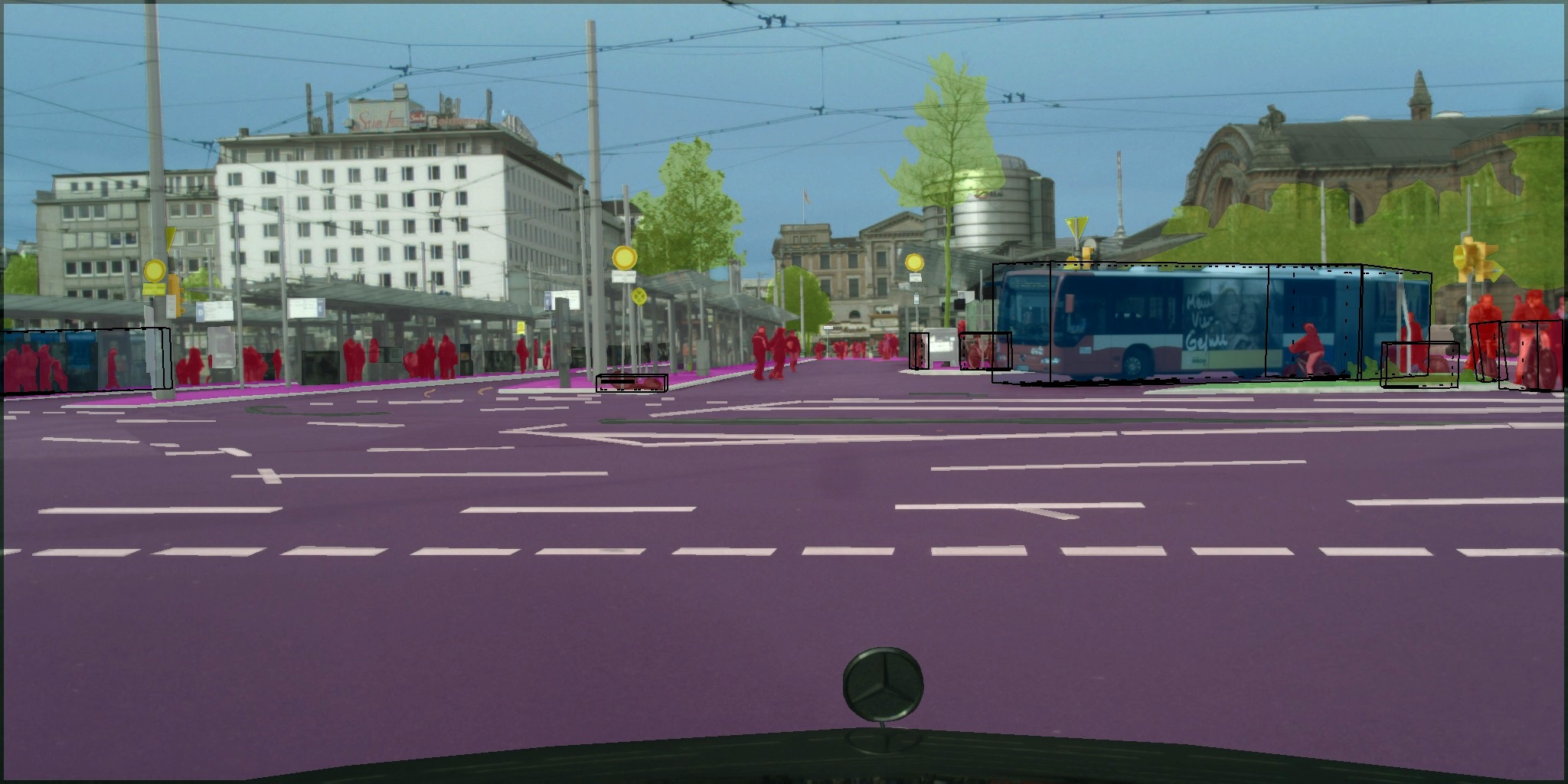}
    \caption{An articulated bus with two distinct 3D bounding boxes.\label{fig:bus}}
\end{figure}

\begin{figure}[t]
    \includegraphics[width=\linewidth]{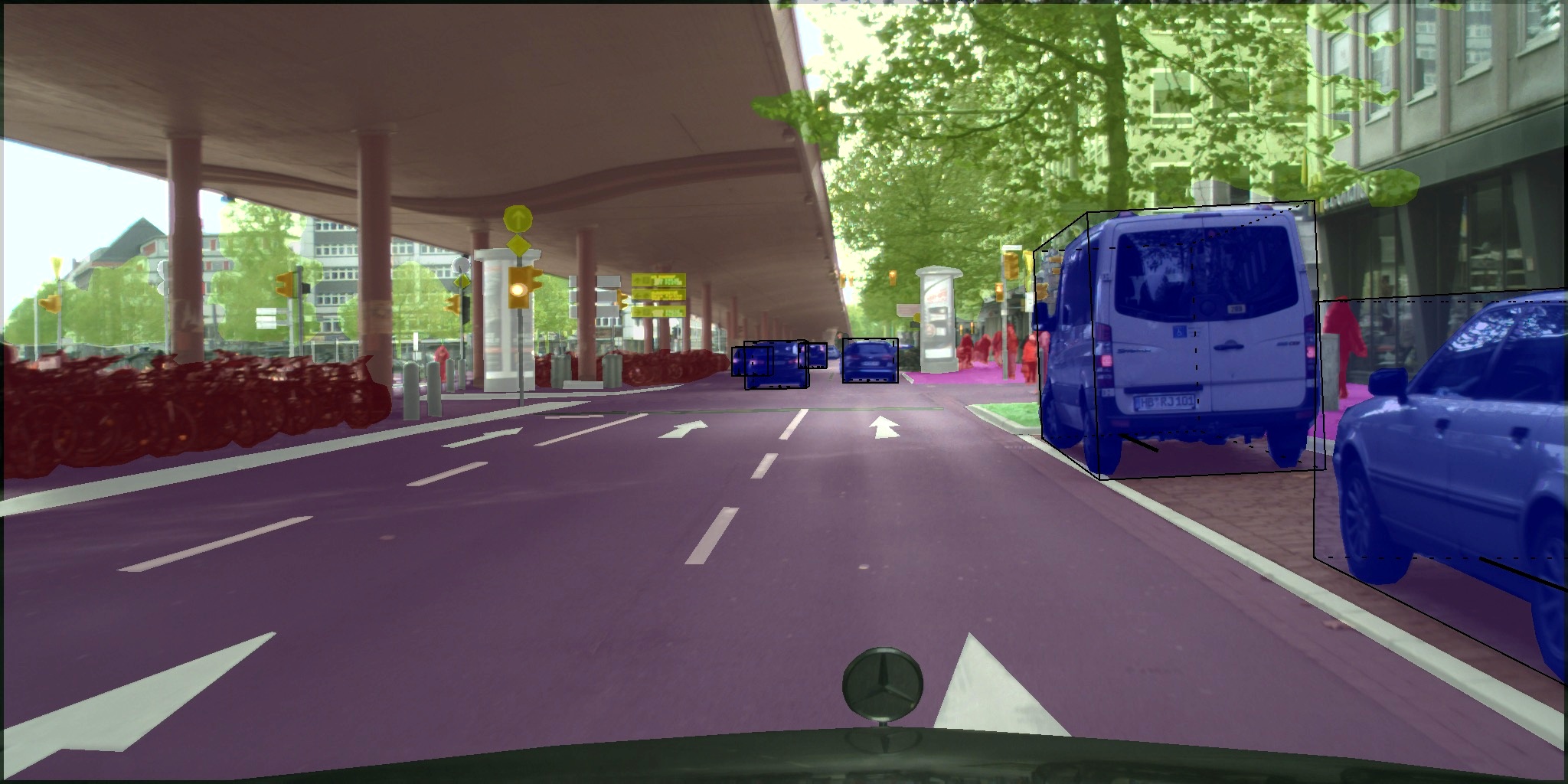}
    \caption{Crowded traffic scene. No single bicycle can be identified. Hence, the whole region is set to \emph{ignore} and overlapping detections will not be considered during the evaluation. \label{fig:ignore}}
\end{figure}

% The 3D bounding box annotations are provided in the camera coordinate frame $C$
% with the origin at the optical center of the left camera in the forward facing
% stereo system. The coordinate frame orientation follows ISO 8855 with $x$,$y$, and $z$
% axis pointing forward, leftward and upward respectively.

% \footnote{Further
% information about calibration and transformation between the Cityscapes
% coordinate frames can be found in the official documentation at
% \url{https://github.com/mcordts/cityscapesScripts/blob/master/docs/csCalibration.pdf}}

In addition to the 3D bounding box annotations, the dataset contains the
mapping between instance segments and 3D bounding box annotations.
Furthermore, we provide the aforementioned metadata including occlusion, truncation,
and selected size prototype per each 3D box.

% Occlusion and truncation are float values between 0 and 1 where 1 indicates
% full occlusion or truncation respectively, \ie the object is not visible at
% all.

% !TeX root = ../egpaper_final.tex

\section{Dataset Analysis}
\label{sec:analysis}
% additional images: with pitch and roll, gelenkbus, group
Cityscapes 3D extends the original dataset \cite{cordts2016cityscapes}, which focuses on
semantic and instance segmentation. The Cityscapes dataset contains \num{5000} images
split into \num{2975} images for training, \num{500} images for validation, and \num{1525} images
for testing.

Our 3D bounding box annotations cover all \num{8} semantic classes in the vehicle
category of the Cityscapes dataset, \ie \emph{car}, \emph{truck}, \emph{bus},
\emph{on rails}, \emph{motorcycle}, \emph{bicycle}, \emph{caravan}, and
\emph{trailer}. Analogously to \cite{cordts2016cityscapes}, we ignore
\emph{caravan} and \emph{trailer} during evaluation. Compared
to other 3D detection datasets, Cityscapes 3D has a high object density, which in
turn indicates complex scenes and hence renders successful detection a challenging task, \cf
\cref{tab:obj_per_img}.
Furthermore, due to the labeling process with size
prototypes as presented in \cref{sec:labeling}, we also enriched Cityscapes
with information about the fine grained types of a vehicle instances, \cf
\cref{fig:vehicletypes} for a statistical analysis.

% !TeX root = ../egpaper_final.tex

\begin{table}[t]
  \caption{Average number of 3D annotations per image for the \emph{train} and \emph{val} sets of different datasets. The column \emph{TBT} combines truck, bus, and train. Cityscapes 3D has a high object density across all classes. \label{tab:obj_per_img}\vspace{3pt}}
  \centering
\begin{tabular}{lcccc}
  \toprule % data from data/X_categories.dat
& Car & TBT & Bicycle & Motorbike\\ \midrule
\hspace{-1mm}ApolloScapes & \textbf{11.6} & {0.0} & {0.0} & {0.0} \\
\hspace{-1mm}Argoverse & {4.1} & {0.3} & {0.1} & {0.001} \\
\hspace{-1mm}KITTI & {4.2} & {0.2} & {0.0} & {0.0} \\
\hspace{-1mm}nuScenes & {3.0} & \textbf{{0.6}} & {0.07} & {0.07} \\
\hspace{-1mm}Waymo & {3.2} & {0.0} & {0.04} & {0.0} \\
%A2D2 & {2.2} & \textbf{{0.6}} & {0.0} & {0.0} \\ \midrule
\hspace{-1mm}Cityscapes 3D & {6.4} & {0.2} & \textbf{{1.2}} & \textbf{{0.2}} \\
\bottomrule \\ 
%\multicolumn{5}{l}{\footnotesize\footnotemark[1] due to its comparable high resolution of \SI{9.2}{MP} and the mounting height of the camera, a relatively large number of vehicles can be resolved in the RGB images.} \\
\end{tabular}
\end{table}
% !TeX root = ../egpaper_final.tex

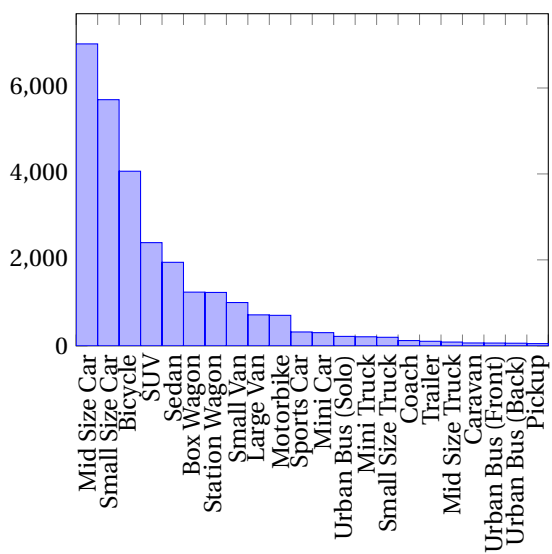
\begin{figure}[t]
	\begin{center}
	\pgfplotstableread[col sep=comma]{prototype_count.dat}\cstable
	\begin{tikzpicture}
		\begin{axis} [
			ybar interval,
			xticklabels from table={\cstable}{Prototyp}, 
			xticklabel style={rotate=90},
			height=6cm,
			grid=none,
			xmin=1,xmax=23,
			ymin=0,
			width=\linewidth,
			title={Vehicle Occurrence},
			tick align=inside
		]
			\addplot table [x=Num, y=Count] from \cstable;
		\end{axis}
	\end{tikzpicture}
	\caption{Number of different vehicle types in Cityscapes 3D \emph{train} and \emph{val}. \label{fig:vehicletypes}}
	\end{center}
\end{figure}
% !TeX root = ../egpaper_final.tex

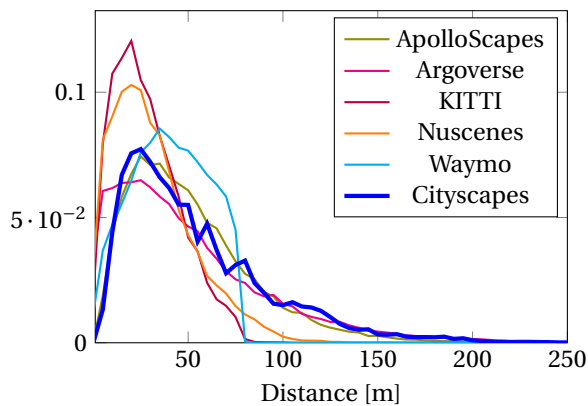
\begin{figure}[t]
	\pgfplotstableread[col sep=comma]{depth_hist.dat}\cstable
	\pgfplotstableread[col sep=comma]{kitti_depth_hist.dat}\kittitable
	\pgfplotstableread[col sep=comma]{nuscenes_depth_hist.dat}\nutable
	\pgfplotstableread[col sep=comma]{apollo_depth_hist.dat}\apollotable
	\pgfplotstableread[col sep=comma]{audi_depth_hist.dat}\auditable
	\pgfplotstableread[col sep=comma]{argo_depth_hist.dat}\argotable
	\pgfplotstableread[col sep=comma]{waymo_depth_hist.dat}\waymotable
	\begin{center}
	\begin{tikzpicture}
	\definecolor{color1}{rgb}{1,0.498039215686275,0.0549019607843137}
		\begin{axis} [
			height=6cm,
			grid=none,
			xmin=1,xmax=250,
			ymin=0,
			width=\linewidth,
			title={Distance Distribution},
			tick align=inside,
			xlabel={Distance [\si{\meter}]},
		]			
		%\addplot+ [mark=none,thick] table [x=Depth, y expr=\thisrow{Prob}*5] from \auditable; 
		%\addlegendentry{A2E2} 

		\addplot+ [ mark=none,thick, olive] table [x=Depth, y expr=\thisrow{Prob}*5] from \apollotable; 
		\addlegendentry{ApolloScapes} 

		\addplot+ [mark=none,thick,magenta] table [x=Depth, y expr=\thisrow{Prob}*5] from \argotable; 
		\addlegendentry{Argoverse}

		\addplot+ [solid, mark=none,thick, purple] table [x=Depth, y expr=\thisrow{Prob}*5] from \kittitable;
		\addlegendentry{KITTI} 

		\addplot+ [mark=none,thick, color1] table [x=Depth, y expr=\thisrow{Prob}*5] from \nutable; 
		\addlegendentry{Nuscenes} 

		\addplot+ [mark=none,thick, cyan] table [x=Depth, y expr=\thisrow{Prob}*5] from \waymotable; 
		\addlegendentry{Waymo} 

		\addplot+ [solid, blue, mark=none,ultra thick] table [x=Depth, y expr=\thisrow{Prob}*5] from \cstable;
		\addlegendentry{Cityscapes}
		
		\end{axis}
	\end{tikzpicture}
	\end{center}
	\caption{Distribution of distance of all objects for state-of-the-art datasets on \emph{train} and \emph{val}. \label{fig:vehicledistances}}
\end{figure}

In contrast to the majority of existing 3D object detection datasets, we use
stereo point clouds to annotate 3D bounding boxes instead of lidar measurements.
This allows us to overcome issues due to the sparsity of lidar measurements, especially
for distant objects. As a result, the distribution of vehicles over the distance
to the camera has a long tail at far distances as
illustrated in \cref{fig:vehicledistances}.

To estimate the quality of our 3D bounding box annotations, we compare
the results of our labeling approach with perfect ground truth from synthetic data.
To this end, we used our annotation tool and workflow to relabel \num{20} images from the
Synscapes dataset \cite{wrenninge2018synscapes}, which was designed to align well
with Cityscapes. The human-labeled annotations are subsequently compared to the
perfect Synscapes ground truth, see \cref{fig:labeling_analysis}.
The analysis shows an average error in the annotated yaw angle of
below \SI{2.1}{\degree} as well as an average center position error of below
\SI{1}{\meter} for all distance levels. Without the bird's-eye view labeling
aid, the average errors in both categories are significantly higher and increase
with the distance of the objects, confirming the effectiveness of our annotation scheme.

%\begin{figure}[t]
%\includegraphics[width=\linewidth]{imgs/stereo_scatter_rgb_frankfurt_294}
%\caption{3D bounding box annotation example with stereo measurements Stereo measurements are shown matched with RGB colors.\label{fig:stereo_labeling}}
%\end{figure}

\begin{figure}[t]
    \newlength\figH
    \newlength\figW
    \setlength{\figH}{\linewidth/2}
    \setlength{\figW}{\linewidth}
    % This file was created by tikzplotlib v0.9.1.
\begin{center}
\begin{tikzpicture}

\definecolor{color0}{rgb}{0.12156862745098,0.466666666666667,0.705882352941177}
\definecolor{color1}{rgb}{1,0.498039215686275,0.0549019607843137}

\begin{groupplot}[group style={group size=1 by 2}]
\nextgroupplot[
axis line style={white!80!black},
height=\figH,
legend cell align={left},
legend style={fill opacity=0.8, draw opacity=1, text opacity=1, at={(0.97,0.03)}, anchor=south east, draw=none},
tick align=outside,
tick pos=left,
width=\figW,
x grid style={white!80!black},
xmajorgrids,
xmin=-0.5, xmax=10.5,
xtick style={color=white!15!black},
xtick={0,1,2,3,4,5,6,7,8,9,10},
xticklabels={10,20,30,40,50,60,70,80,90,100,Inf},
y grid style={white!80!black},
ylabel={Yaw Error [\si{\degree}]},
ymajorgrids,
ymin=0, ymax=4,
ytick style={color=white!15!black}
]
\addplot [semithick, color0]
table {%
0 1.78
1 1.18
2 1.16
3 1.95
4 2.06
5 2.06
6 2.08
7 2.07
8 2.07
9 2.07
10 2.09
};
\addlegendentry{w/ Stereo BEV}
\addplot [semithick, color1]
table {%
0 1.52
1 1.84
2 2.62
3 3.12
4 3.1
5 3.3
6 3.62
7 3.63
8 3.6
9 3.53
10 3.65
};
\addlegendentry{wo/ Stereo BEV}

\nextgroupplot[
axis line style={white!80!black},
height=\figH,
tick align=outside,
tick pos=left,
width=\figW,
x grid style={white!80!black},
xlabel={Distance up to [\si{\meter}]},
xmajorgrids,
xmin=-0.5, xmax=10.5,
xtick style={color=white!15!black},
xtick={0,1,2,3,4,5,6,7,8,9,10},
xticklabels={10,20,30,40,50,60,70,80,90,100,Inf},
y grid style={white!80!black},
ylabel={Center Error [\si{\meter}]},
ymajorgrids,
ymin=0, ymax=4,
ytick style={color=white!15!black}
]
\addplot [semithick, color0]
table {%
0 0.61
1 0.64
2 0.68
3 0.81
4 0.89
5 0.93
6 0.93
7 0.94
8 0.94
9 0.94
10 0.75
};
\addplot [semithick, color1]
table {%
0 0.86
1 1.21
2 1.46
3 1.81
4 1.97
5 2.24
6 2.48
7 2.62
8 2.69
9 3.01
10 3.64
};
\end{groupplot}

\end{tikzpicture}
\end{center}
    \caption{Labeling quality evaluation of \num{20} relabeled images in the Synscapes \cite{wrenninge2018synscapes} dataset using the Cityscapes 3D labeling workflow. Labelers were asked to label 3D bounding boxes w/ and w/o the stereo bird's-eye view. Annotation quality is compared to the synthetic ground truth.\label{fig:labeling_analysis}}
\end{figure}
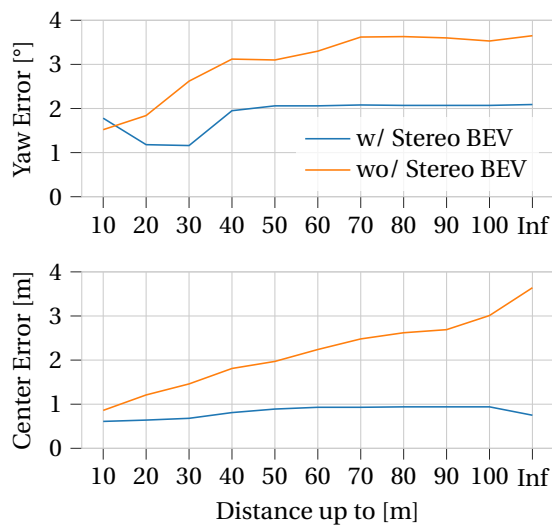

% !TeX root = ../egpaper_final.tex

\section{Benchmark Metrics}
\label{sec:benchmark}

To evaluate the performance of an RGB-based detection model in the Cityscapes 3D benchmark, we
propose several metrics in the following that individually assess recognition performance in terms of 2D and 3D detection and localization.
All metrics are then combined into a single detection score per class.
The mean over all classes is denoted as mean
detection score (mDS) and is used for ranking the approaches within the benchmark.
As described in \cref{sec:analysis}, the \num{8} vehicle classes of the Cityscapes instance
segmentation benchmark are evaluated taking ignore regions into account, \cf \cref{sec:labeling}.
The circumscribing 2D bounding box of an ignore region defines this region as \emph{ignore}
and overlapping detections within these regions will not count as false positive.

The detection performance in popular benchmarks is commonly evaluated
either distance independent or only for very coarse distance intervals.
KITTI \cite{geiger2012we} and nuScenes \cite{nuscenes2019} use the same
detection thresholds for all distances and hence count all objects equally, irrespective
of their distance to the camera.
The metrics of Waymo Open \cite{waymo_open_dataset} cluster objects in three rather coarse ranges from \num{0} to
\SI{35}{\meter}, from \num{35} to \SI{50}{\meter} and from \SI{50}{\meter} to inf.
In contrast, we aim for a more detailed depth-dependent performance analysis.

As our benchmark focuses on detecting 3D bounding boxes from monocular RGB
images, we employ a metric consisting of two factors.
The first factor is the 2D Average Precision (AP) metric, which is
well-known and the standard metric to assess 2D bounding box detection performance.

As second factor, we summarize three-dimensional core properties, \ie center position
in 3D space, orientation, and size. These aspects are evaluated individually using
Depth-Dependent True Positive (DDTP) metrics. These metrics are
only calculated for true positive detections and their corresponding ground truth counterpart.
The 3D ground truth annotations are binned based on their distance to the camera such
that we can obtain a depth-dependent analysis.
For each bin and DDTP metric, we compute the average score in terms of accuracy of
the underlying 3D property.
Eventually, the DDTP value is determined as the average over all distance bins.

This two-factor approach allows to distinguish between semantics and geometry.
For example, a detection with an accurate projection into the RGB image
but with a poor distance estimation counts as a true positive in the
2D AP computation and at the same time lowers the DDTP metric that
assesses the quality of 3D localization. In contrast, the metric definitions of \eg KITTI \cite{geiger2012we} or
nuScenes \cite{nuscenes2019} do not provide such a fine analysis and lead to an
overall decreased model accuracy.

% AP as well as DDTP metrics can be visualized as shown in \cref{fig:spider}.

In the following $X_\text{max}$ is defined as the maximum detection radius up to
which the object detection performance shall be analyzed. A bin $D(s, c)$ denotes the
set of pairs of detections $d$ and ground truth objects $g$ within the depth range $[s, s + \delta_s)$
and a minimum detection confidence of $c$. $N$ is the cardinality of $D(s, c)$.
We set the maximum detection depth $X_\text{max} = \SI{100}{\meter}$ as \SI{90}{\percent}
of all annotated objects are within \SI{100}{\meter} distance to the ego-vehicle.
We choose $\delta_s = \SI{5}{\meter}$ for the bin size.

\subsection{2D Average Precision (AP)}
We use standard Average Precision (AP) \cite{everingham2010pascal}
for assessing the 2D detection performance based on 2D bounding boxes.
The 2D bounding box of both, ground truth and predicted 3D box,
is defined as the circumscribing rectangle of all 8 vertices of the
3D box projected into the image.
Matching is conducted in the 2D space and we require an IoU of $0.7$ between ground truth and
detection to accept the detection as true positive.

Detections for which at least \SI{70}{\percent} of the predicted 2D bounding box
covers any ignore region will be discarded and not included in the evaluation.
Ignore regions typically contain groups of objects that
cannot be visually separated or objects with high occlusion or truncation, \cf \cref{sec:labeling}.

Overall, AP is defined as
\begin{equation}
	\text{AP }= \int p(r) \mathrm dr
	\label{eq:2dap}
\end{equation}
with $p(r)$ being the precision value for recall $r$.

\subsection{Depth-Dependent Average Precision}
We calculate standard AP for all objects within the range $[s, s+ \delta_s)$. To
assign a depth value to an object, we take the ground truth depth for true positive and false
negative detections and the predicted depth for false positives.

\subsection{Depth-Dependent True Positive Metrics}
We calculate several depth-dependent true positive metrics, \ie \emph{BEV
	Center Distance}, \emph{Yaw Similarity}, \emph{Pitch-Roll Similarity}, and
\emph{Size Similarity}. The depth of the ground truth box is used to determine the
applicable interval $[s, s + \delta_s)$. In contrast to the regular 2D AP score \cref{eq:2dap},
we use a fixed confidence threshold for the depth-dependent true positive metrics.
The threshold $c_w$ is defined as
\begin{equation}
	c_w = \underset{c \in [0,1]}{\operatorname{arg\, max}} p(c)r(c)
\end{equation}
with $p(c)$ and $r(c)$ denoting the precision and the recall for score $c$.
Fixing the confidence to $c_w$ is motivated by the observation that most
perception systems use fixed confidence thresholds to limit the number of
detected objects and hence requirements and bandwidths for downstream
applications. By this definition we allow for an assessment of
the quality of each model \emph{as is} when deployed.

\paragraph{Center Distance}
Bird's-Eye View Center Distance is defined as the normalized integral of the
depth-dependent distance up to the maximum depth of interest $X_\text{max}$, \ie
\begin{equation}
	\text{BEVCD} = 1 - \frac 1 {X_\text{max}^2} \int _ 0 ^{X_\text{max}}k(s) \mathrm ds  \label{eq:bevcd}
\end{equation}
with
\begin{equation}
	k(s)         = \frac 1 N  \sum _{ d, g \in D(s,c_w)} \min \left(X_\text{max} , \sqrt {\sum_{i \in \lbrace x,y\rbrace} (d_i - g_i)^2}\right) .
\end{equation}
As the maximal center distance is limited to $X_\text{max}$ the integral is
scaled by the inverse of $X_\text{max}^2$ to obtain a value between $0$ and $1$.

\paragraph{Yaw Similarity}
Following the same scheme, Yaw Similarity is inspired by \cite{geiger2012we} and
is defined as
\begin{equation}
	\text{YawSim} =\frac 1 {X_\text{max}} \int _ 0 ^{X_\text{max}}k(s) \mathrm ds  \label{eq:bevcd}
\end{equation}
with
\begin{equation}
	k(s)          = \frac 1 N  \sum _{ d, g \in D(s,c_w)} \frac{1 + \cos{\left( \Delta_\text{Yaw} \right)}}{2}.
\end{equation}
However, it is not required to scale by the squared value of $X_\text{max}$ as
$\frac {1 + \cos}2 $ is limited between $0$ and $1$.

\paragraph{Pitch-Roll Similarity}
Pitch-Roll Similarity is calculated analogously to Yaw Similarity but both pitch
and roll orientation are combined since pitch and roll of a vehicle are not
independent in realistic driving scenarios, \ie
\begin{equation}
	\text{PRSim} =\frac 1 {X_\text{max}} \int _ 0 ^{X_\text{max}}k(s) \mathrm ds  \label{eq:bevcd}
\end{equation}
with
\begin{equation}
	k(s)         = \frac 1 N  \sum _{ d, g \in D(s,c_w)} \frac{2 + \cos{\left( \Delta_\text{Pitch} \right)} + \cos{\left( \Delta_\text{Roll} \right)}}4 .
\end{equation}

\paragraph{Size Similarity}
Size Similarity assesses the 3D dimensions of the true positive detection and is defined as
\begin{equation}
	\text{SizeSim} =\frac 1 {X_\text{max}} \int _ 0 ^{X_\text{max}}k(s) \mathrm ds  \label{eq:bevcd}
\end{equation}
with
\begin{equation}
	k(s)           = \frac 1 N  \sum _{ d, g \in D(s,c_w)}\prod _{x \in \lbrace l,w,h \rbrace} \min \left( \frac {d_x}{g_x}, \frac {g_x}{d_x}\right)
\end{equation}
with $l,w,h$ being length, width, and height of the corresponding 3D bounding box.

\subsection{Detection Score}
To combine all quality measures, we define the detection score per class as
\begin{equation}
	\text{DS} =  \text{AP} \times \frac { \text{BEVCD} + \text{YawSim} + \text{PRSim} + \text{SizeSim}} 4.
\end{equation}
By this definition, we enforce the 2D Average Precision to be an upper bound for
the final detection score that can only be reached if all true positive bounding boxes are
predicted perfectly. Furthermore, depth-dependent AP does not count to the
overall detection score. Finally, the overall detection score mDS is calculated
as the mean of all detection scores per class. mDS is consequently used for the benchmark ranking.

%\begin{figure}
%	\includegraphics[width=1.07\linewidth]{chapters/kiviat_sample.tikz}
%	\caption{Visualization of the calculated metrics for two classes. Each DDTP metric is scaled by AP. \label{fig:spider}}
%\end{figure}

\section{Conclusion}
In this work, we presented a novel extension to the popular Cityscapes dataset,
enriching the annotations with high quality 3D bounding boxes for vehicles. With
this extension, we specifically aim at motivating progress in the important
research area of monocular 3D object detection for autonomous driving.

We identified two major shortcomings of current state-of-the-art 3D object
detection datasets and corresponding benchmarks that we address: (i) The
majority of existing 3D object detection datasets relies on lidar point clouds
for labeling. This introduces errors when projecting the annotations into
RGB images if the cross-sensor calibration or synchronization are
imperfect, thus hindering effective benchmarking of monocular 3D object
detection methods. We address this issue by labeling the 3D box annotations
using only RGB and stereo information, independent from multi-sensor calibration or
synchronization. Hence, our 3D bounding box annotations are consistent in both,
image and 3D space.
(ii) Benchmark metrics of existing 3D object
detection datasets are often relying on a minimum 3D IoU overlap for true
positive detections which can be extremely hard to achieve with monocular
detection methods, favoring lidar-based detection algorithms. Furthermore,
state-of-the-art metrics are usually distance agnostic.
However, especially in autonomous driving, the distance of objects to
the ego-vehicle strongly correlates with the relevance of an object for the actual driving task.
We address these findings by using 2D IoU thresholds for true positive detections,
which adapts the matching optimally for monocular detection methods.
Furthermore, we provide a set of novel, distance-dependent metrics that enable
the benchmarking of monocular 3D object detection methods
for autonomous driving.

{\small
    \bibliographystyle{ieee_fullname}
    \bibliography{egbib}

\begin{thebibliography}{10}\itemsep=-1pt

\bibitem{waymo_open_dataset}
Waymo open dataset: An autonomous driving dataset.
\newblock \url{https://waymo.com/open/}, 2019.

\bibitem{bao2019monofenet}
Wentao Bao, Bin Xu, and Zhenzhong Chen.
\newblock Monofenet: Monocular 3d object detection with feature enhancement
  networks.
\newblock {\em IEEE Transactions on Image Processing}, 2019.

\bibitem{boxy2019}
Karsten Behrendt.
\newblock Boxy vehicle detection in large images.
\newblock In {\em ICCV}, 2019.

\bibitem{brazil2019m3d}
Garrick Brazil and Xiaoming Liu.
\newblock M3d-rpn: Monocular 3d region proposal network for object detection.
\newblock In {\em ICCV}, 2019.

\bibitem{cabon2020vkitti2}
Yohann Cabon, Naila Murray, and Martin Humenberger.
\newblock Virtual kitti 2.
\newblock {\em arXiv preprint arXiv:2001.10773}, 2020.

\bibitem{nuscenes2019}
Holger Caesar, Varun Bankiti, Alex~H. Lang, Sourabh Vora, Venice~Erin Liong,
  Qiang Xu, Anush Krishnan, Yu Pan, Giancarlo Baldan, and Oscar Beijbom.
\newblock nuscenes: A multimodal dataset for autonomous driving.
\newblock {\em arXiv preprint arXiv:1903.11027}, 2019.

\bibitem{Argoverse}
Ming-Fang Chang, John~W Lambert, Patsorn Sangkloy, Jagjeet Singh, Slawomir Bak,
  Andrew Hartnett, De Wang, Peter Carr, Simon Lucey, Deva Ramanan, and James
  Hays.
\newblock Argoverse: 3d tracking and forecasting with rich maps.
\newblock In {\em CVPR}, 2019.

\bibitem{cordts2016cityscapes}
Marius Cordts, Mohamed Omran, Sebastian Ramos, Timo Rehfeld, Markus Enzweiler,
  Rodrigo Benenson, Uwe Franke, Stefan Roth, and Bernt Schiele.
\newblock The cityscapes dataset for semantic urban scene understanding.
\newblock In {\em CVPR}, 2016.

\bibitem{everingham2010pascal}
Mark Everingham, Luc Van~Gool, Christopher~KI Williams, John Winn, and Andrew
  Zisserman.
\newblock The pascal visual object classes (voc) challenge.
\newblock {\em International journal of computer vision}, 88(2):303--338, 2010.

\bibitem{gaehlert2019vg}
Nils G{\"{a}}hlert, Niklas Hanselmann, Uwe Franke, and Joachim Denzler.
\newblock Visibility guided nms: Efficient boosting of amodal object detection
  in crowded traffic scenes.
\newblock In {\em NeurIPS Workshops}, 2019.

\bibitem{gaidon2016virtual}
Adrien Gaidon, Qiao Wang, Yohann Cabon, and Eleonora Vig.
\newblock Virtual worlds as proxy for multi-object tracking analysis.
\newblock In {\em CVPR}, 2016.

\bibitem{geiger2012we}
Andreas Geiger, Philip Lenz, and Raquel Urtasun.
\newblock Are we ready for autonomous driving? the kitti vision benchmark
  suite.
\newblock In {\em CVPR}, 2012.

\bibitem{aev2019}
Jakob Geyer, Yohannes Kassahun, Mentar Mahmudi, Xavier Ricou, Rupesh Durgesh,
  Andrew~S. Chung, Lorenz Hauswald, Viet~Hoang Pham, Maximilian Mühlegg,
  Sebastian Dorn, Tiffany Fernandez, Martin Jänicke, Sudesh Mirashi,
  Chiragkumar Savani, Martin Sturm, Oleksandr Vorobiov, Martin Oelker,
  Sebastian Garreis, and Peter Schuberth.
\newblock {A2D2: AEV Autonomous Driving Dataset}.
\newblock \url{http://www.a2d2.audi}, 2019.

\bibitem{hirschmuller2007stereo}
Heiko Hirschm{\"u}ller.
\newblock Stereo processing by semiglobal matching and mutual information.
\newblock {\em TPAMI}, 30(2):328--341, 2007.

\bibitem{huang2018apolloscape}
Xinyu Huang, Xinjing Cheng, Qichuan Geng, Binbin Cao, Dingfu Zhou, Peng Wang,
  Yuanqing Lin, and Ruigang Yang.
\newblock The apolloscape dataset for autonomous driving.
\newblock In {\em CVPR Workshops}, 2018.

\bibitem{lyft2019}
R. Kesten, M. Usman, J. Houston, T. Pandya, K. Nadhamuni, A. Ferreira, M. Yuan,
  B. Low, A. Jain, P. Ondruska, S. Omari, S. Shah, A. Kulkarni, A. Kazakova, C.
  Tao, L. Platinsky, W. Jiang, and V. Shet.
\newblock Lyft level 5 av dataset 2019.
\newblock \url{https://level5.lyft.com/dataset/}, 2019.

\bibitem{ku2019monocular}
Jason Ku, Alex~D Pon, and Steven~L Waslander.
\newblock Monocular 3d object detection leveraging accurate proposals and shape
  reconstruction.
\newblock In {\em CVPR}, 2019.

\bibitem{li2019gs3d}
Buyu Li, Wanli Ouyang, Lu Sheng, Xingyu Zeng, and Xiaogang Wang.
\newblock Gs3d: An efficient 3d object detection framework for autonomous
  driving.
\newblock In {\em CVPR}, 2019.

\bibitem{liu2019deep}
Lijie Liu, Jiwen Lu, Chunjing Xu, Qi Tian, and Jie Zhou.
\newblock Deep fitting degree scoring network for monocular 3d object
  detection.
\newblock In {\em CVPR}, 2019.

\bibitem{ma2019accurate}
Xinzhu Ma, Zhihui Wang, Haojie Li, Pengbo Zhang, Wanli Ouyang, and Xin Fan.
\newblock Accurate monocular 3d object detection via color-embedded 3d
  reconstruction for autonomous driving.
\newblock In {\em ICCV}, 2019.

\bibitem{manhardt2019roi}
Fabian Manhardt, Wadim Kehl, and Adrien Gaidon.
\newblock Roi-10d: Monocular lifting of 2d detection to 6d pose and metric
  shape.
\newblock In {\em CVPR}, 2019.

\bibitem{neuhold2017mapillary}
Gerhard Neuhold, Tobias Ollmann, Samuel Rota~Bulo, and Peter Kontschieder.
\newblock The mapillary vistas dataset for semantic understanding of street
  scenes.
\newblock In {\em ICCV}, 2017.

\bibitem{pham20193d}
Quang-Hieu Pham, Pierre Sevestre, Ramanpreet~Singh Pahwa, Huijing Zhan, Chun~Ho
  Pang, Yuda Chen, Armin Mustafa, Vijay Chandrasekhar, and Jie Lin.
\newblock A* 3d dataset: Towards autonomous driving in challenging
  environments.
\newblock {\em arXiv preprint arXiv:1909.07541}, 2019.

\bibitem{richter2017playing}
Stephan~R Richter, Zeeshan Hayder, and Vladlen Koltun.
\newblock Playing for benchmarks.
\newblock In {\em ICCV}, 2017.

\bibitem{ros2016synthia}
German Ros, Laura Sellart, Joanna Materzynska, David Vazquez, and Antonio~M
  Lopez.
\newblock The synthia dataset: A large collection of synthetic images for
  semantic segmentation of urban scenes.
\newblock In {\em CVPR}, 2016.

\bibitem{simonelli2019disentangling}
Andrea Simonelli, Samuel~Rota Bulo, Lorenzo Porzi, Manuel L{\'o}pez-Antequera,
  and Peter Kontschieder.
\newblock Disentangling monocular 3d object detection.
\newblock In {\em ICCV}, 2019.

\bibitem{wrenninge2018synscapes}
Magnus Wrenninge and Jonas Unger.
\newblock Synscapes: A photorealistic synthetic dataset for street scene
  parsing.
\newblock {\em arXiv preprint arXiv:1810.08705}, 2018.

\bibitem{yang2019std}
Zetong Yang, Yanan Sun, Shu Liu, Xiaoyong Shen, and Jiaya Jia.
\newblock Std: Sparse-to-dense 3d object detector for point cloud.
\newblock In {\em ICCV}, 2019.

\bibitem{ye2020sarpnet}
Yangyang Ye, Houjin Chen, Chi Zhang, Xiaoli Hao, and Zhaoxiang Zhang.
\newblock Sarpnet: Shape attention regional proposal network for lidar-based 3d
  object detection.
\newblock {\em Neurocomputing}, 379:53--63, 2020.

\bibitem{yu2018bdd100k}
Fisher Yu, Wenqi Xian, Yingying Chen, Fangchen Liu, Mike Liao, Vashisht
  Madhavan, and Trevor Darrell.
\newblock Bdd100k: A diverse driving video database with scalable annotation
  tooling.
\newblock {\em arXiv preprint arXiv:1805.04687}, 2018.

\end{thebibliography}
}

\end{document}